\pdfoutput=1

\documentclass[11pt]{article}

\usepackage[]{ACL2023}

\usepackage{graphicx}

\usepackage{times}
\usepackage{latexsym}
\usepackage{balance}
\usepackage{enumitem}
\usepackage[T1]{fontenc}

\usepackage[utf8]{inputenc}

\usepackage{microtype}

\usepackage{inconsolata}

\newcommand{\sds}[0]{\textsc{sds}}

\newcommand{\hri}[0]{\textsc{hri}}

\newcommand{\nlp}[0]{\textsc{nlp}}
\newcommand{\llm}[0]{\textsc{llm}}
\newcommand{\mmllm}[0]{\textsc{mm-llm}}
\newcommand{\rlhf}[0]{\textsc{rlhf}}

\title{Dialogue with Robots: Proposals for Broadening Participation \\ and Research in the SLIVAR Community}

\author{Casey Kennington\thanks{\ Authors Kennington, Alikhani, and Pon-Barry proposed and organized the workshop. The remaining authors are listed alphabetically. Corresponding author:\\ \url{caseykennington@boisestate.edu}} \\ Boise State \\University \And
Malihe Alikhani$^*$ \\ Northeastern \\ University \And
Heather Pon-Barry$^*$ \\ Mount Holyoke \\ College \And
Katherine Atwell \\ University of \\ Pittsburgh \AND 
Yonatan Bisk \\ Carnegie Mellon \\ University \And 
Daniel Fried \\ Carnegie Mellon \\ University \And
Felix Gervits \\ US Army \And
Zhao Han \\ University of \\ South Florida \AND
Mert Inan \\ Northeastern \\ University \And
Michael Johnston \\ Amazon \And
Raj Korpan \\ Hunter College, \\ City University of \\ New York \And
Diane Litman \\ University of \\ Pittsburgh \AND
Matthew Marge \\ DARPA \And
Cynthia Matuszek \\ University of Maryland, \\ Baltimore County \And
Ross Mead \\ Semio \And
Shiwali Mohan \\ Xerox PARC \AND
Raymond Mooney \\ University of Texas, \\ Austin \And
Natalie Parde \\ University of Illinois, \\ Chicago \And
Jivko Sinapov \\ Tufts \\ University \And
Angela Stewart \\ University of \\ Pittsburgh \AND
Matthew Stone \\ Rutgers \\ University\And
Stefanie Tellex \\ Brown \\ University \And
Tom Williams \\  Colorado \\ School of Mines
}

\begin{document}
\maketitle

\begin{abstract}
The ability to interact with machines using natural human language is becoming not just commonplace, but expected. The next step is not just text interfaces, but speech interfaces and not just with computers, but with all machines including robots. In this paper, we chronicle the recent history of this growing field of spoken dialogue with robots and offer the community three proposals, the first focused on education, the second on benchmarks, and the third on the modeling of language when it comes to spoken interaction with robots. The three proposals should act as white papers for any researcher to take and build upon. 

\end{abstract}

\section{Introduction}

Daily use of technology that allows humans to interact with machines using natural language by the general population is accelerating. The field of Natural Language Processing (\nlp) was transformed by transformers \cite{Vaswani2017-kv} since BERT \cite{Devlin2018-lz}, and more recent scaling of transformer language models like ChatGPT has brought what used to be models only known in the \nlp\ research community to consumer use where generating language is a valuable outcome, for example in generating drafts of emails or company policies. However, language use is not just confined to text; much of human interaction is still face-to-face where spoken language is used to refer to objects and events in the world. For example, cooking, yard work, travel, construction and many other domains of every day life require humans to interact with physical objects that large language models like ChatGPT can talk about generally, but not refer to specifically. Robots, in contrast, exist in a physical embodiment and are being used in cooking, yard work, travel, and construction to assist humans, automate mundane tasks such as vacuuming, perform tasks that are dangerous to humans, and act as socially aware and interactive agents for people who are lonely. 

\citet{Kennington2021-qh} explained that a newly forming \emph{Special Interest Group for Spoken Language Interaction with Virtual Agents and Robots} (SLIVAR) aims to bring the broad areas of spoken dialogue systems, robotics, and human-robot interaction together with the goal of empowering people to communicate with robots the way that humans largely communicate with each other: natural human language, particularly spoken dialogue. A \emph{robot} is an actuated mechanism programmable in two or more axes with a degree of autonomy, moving within its environment, to perform intended tasks that serve the needs of a user \cite{Eskenazi2020-no}.\footnote{This, as well as other arguments and definitions that follow are unceremoniously lifted from Roger Moore's excellent keynote at RO-MAN 2020  (\url{https://tinyurl.com/rmoreroman})}$^,$\footnote{See also this Twitter thread: \url{https://twitter.com/BLeichtmann/status/1314080122169970688}}  A \emph{spoken dialogue system} (\sds) is a an automated system that is able to converse with a human with voice.\footnote{Here, the focus is on robots and speech, though in many cases the arguments made here also extend to virtual agents and communication using a text chat medium.}

Recent workshops and events have facilitated the convergence of dialogue interaction with robots. Two U.S. National Science Foundation (NSF) workshops were convened in October, 2019. First, the \emph{Future Directions Workshop, Toward User-Oriented Agents: Research directions and Challenges} \cite{Eskenazi2020-no} which focused on the role of intelligent agents and how to make them more user-oriented. The participants of the workshop identified broad areas and themes for future directions including applications, infrastructure, dynamic views of user-agent interaction, and made several recommendations in building low-cost dialogue systems, multimodal, grounded, and situated interaction, robust and flexible dialogue management, and intelligent agents as good actors. The second workshop, \emph{Spoken Language Interaction with Robots} \cite{Marge2022-yf}, focused on speech and the complexities thereof when robots are involved.

A Dagstuhl (Germany) Seminar also convened in January 2020 on the topic of SLIVAR.\footnote{Dagstuhl ID 20021 \url{https://drops.dagstuhl.de/opus/volltexte/2020/12400/}} This resulted in organization of other events such as a special session on robots and dialogue (RoboDial 2.0) at the SIGDIAL 2020 conference,\footnote{\url{https://robodial.github.io}} a workshop on natural language generation at the HRI 2020 conference,\footnote{\url{https://hbuschme.github.io/nlg-hri-workshop-2020/}} and a workshop ROBOTDIAL at the IJCAI 2020 conference.\footnote{\url{http://sap.ist.i.kyoto-u.ac.jp/ijcai2020/robotdial/}} The primary goal of the Dagstuhl Seminar was to provide discussion and establish a community. Discussions revolved around ethics, usability, scenarios for human-agent / human-robot groups, evaluation, architectures, and situated language understanding. Other recent, related events include a AAAI Symposium on Natural Communication for Human-Robot Collaboration (2017),\footnote{\url{https://www.ttic.edu/nchrc/}} and RoboNLP workshops including spatial language understanding.\footnote{See links to prior workshops \url{https://splu-robonlp.github.io/}}

More recently, on 7-8 April 2023, a workshop funded by NSF convened to discuss ten important topics that follow directly from the prior workshops and ongoing research trends:

\begin{itemize}[leftmargin=*]
    \item Research platforms for robots+spoken dialogue systems
    \item Collection of situated and multimodal corpora
    \item Multimodality: which modalities (beyond vision) are necessary, and how to capture and leverage them
    \item Natural language understanding in co-located, situated, multimodal systems
    \item Natural language generation in co-located, situated, multimodal systems
    \item Representation of the state of the world, robot, and human interlocutor, and building of common ground
    \item Dialogue with robots: clarification, turn-taking, handling ambiguity
    \item Identifying ethical issues surrounding human-robot dialogue
    \item What should a robotics researcher know when they want to add the ability to talk to their robots, and visa-versa?
    \item Identify and plan tasks and benchmarks
\end{itemize}

Furthermore, the workshop included keynote addresses by Yonatan Bisk (CMU) and Stefanie Tellex (Brown University) on the broad topics of robots, dialogue, and vision about where the field is headed given the current state of technology advancement.

The purpose of this paper is to report on that workshop in the form of 3 overarching proposals with the goal to build the SLIVAR community by (1) providing educational resources, (2) establishing benchmarks and challenges, and (3) integrating large language models effectively with robots while meeting important requirements for natural interaction. In the subsequent sections below, each proposal is explained. We then conclude. 

\section{Proposal 1: Educational Resources}

Robotics, natural language processing (\nlp), spoken dialog systems (\sds), and human-robot interaction (\hri) are separate fields, each requiring a lot of educational preparation before a student is ready to perform research ranging from mechanical and electrical engineering to computer and data science as well as machine learning. There are many technical skills for each, ranging from understanding hardware and software to social interactions. The breadth of topics is daunting, yet ample depth in each area is necessary for effective research. 

The depth that each student needs in each area is partially dependent on the kinds of research that the student will engage in. If the student is more interested, for example, in language/symbol grounding, then more coursework that enables a deeper understanding of semantic theories, as well as in-depth understanding of \nlp\ and computer vision is crucial. Whereas, a student more interested in the social aspects of interaction between humans and robots may not need to take courses where they learn how to build robots or worry about navigation, though they will need to be able to build systems and work with various sensor information. 

\subsection{Proposal}

We propose the creation of a central resource that allows members of the community to share syllabi, course content, lecture slides, example code, assignments, assessment methods, platforms, and research tools used in coursework. Much of the coursework in certain areas like Math and Computer Science is likely already covered at many institutions of higher education, but courses in more specific areas like \nlp, \sds, \hri, and robotics might need supplementation for students where those courses are not available.  

\subsubsection{Existing Work}

Courses that have public content exist that could be adopted by others who seek to incorporate relevant coursework to their curricula, including Math, Computer Science, and Machine Learning courses. More specific to the area of interest here are existing courses (linked in the repository above) like Grounding Natural Language, Talking to Robots, Multimodal Machine Learning, Grounded NLP, NLP with a bent on computational semantics including symbol grounding \cite{Kennington2021-uj}, Probabilistic Robots for Human-Robot Interaction, Language and Vision, and Dialogue Systems. Some robot platforms such as GoPiGo\footnote{\url{https://gopigo.io}} and DuckieTown\footnote{\url{https://www.duckietown.org}} offer educational resources with their robot platforms.

Below we outline a proposed sequence of courses for a student to take, and offer some ideas as to how this effort could be organized. 

\subsubsection{Coursework}

Below is a sample of known courses that could be offered in an undergraduate curriculum that would prepare a student to do research at the intersection of robotics, \sds, and \hri. Each bullet point can either be an entire course, or crucial topics to be taught within a course.

\begin{itemize}[noitemsep,leftmargin=*]
    \item Math
    \begin{itemize}
        \item Calculus
        \item Linear Algebra
        \item Statistics
        \item Probability and Information Theory
        \item Differential Equations
    \end{itemize}
    \item Computer Science
    \begin{itemize}
        \item Data Structures
        \item Algorithms
        \item Distributed Systems
    \end{itemize}
    \item Robotics
    \begin{itemize}
        \item Morphology \& Action Planning
        \item Localization \& Navigation
    \end{itemize}
    \item Data Science and Machine Learning
    \begin{itemize}
        \item Data Science (data analysis, munging, visualization)
        \item Machine Learning
        \item Deep Learning
    \end{itemize}
    \item Artificial Intelligence
    \begin{itemize}
        \item Natural Language Processing (large language models)
        \item Computer Vision and Grounding (multimodal machine learning)
        \item Spoken Dialogue Systems (incremental, multimodal)        
    \end{itemize}    
    
    \item Human-Robot Interaction
    \begin{itemize}
        \item Human-Robot Interaction
        \item Scientific Methods (human subjects, evaluation)
    \end{itemize}

\item Ethics and Society
    \begin{itemize}
        \item Measuring Harm in Human-Robot Dialogue
        \item Bias 
        \item Efficiency and Environmental Cost
        \item Accessibility 
    \end{itemize}
    
\end{itemize}

\subsection{Sharing Resources} We propose the creation of a repository of resources for education of students at this crucial, yet complex intersection between \hri, \sds, and robotics. To make it easy to add and share resources, we propose using a GitHub repository where the main landing page has links to the resources. Contributors can use git functionality like pull requests to add information. We have started the \url{https://github.com/bsu-slim/slivar-resources/} repository to serve this purpose. We also propose some kind of forum for discussion about resources such as Discord that anyone can join.\footnote{Contact the corresponding author if you want to join the Discord server.}

\section{Proposal 2: Benchmarks \& Challenges}

Benchmarks and challenges help research move forward by giving a community a common way to compare their work. For example, the GLUE benchmark is a well-known suite of tasks in \nlp\ that has been widely used; the 2018 paper has over 4,000 citations \cite{Wang2018-jg}. However, benchmarks have their drawbacks. \citet{Bowman2021-xn} explained how GLUE and other language understanding benchmarks are not guaranteed to properly test models (indeed, even models that perform well on benchmarks fail at simple language phenomena). 
According to the authors, benchmarks should:

\begin{enumerate}[noitemsep]
    \item Offer a valid test of the relevant language phenomena
    \item Be built around consistently labeled data
    \item Offer adequate statistical power
    \item Disincentivize the use of systems with potentially harmful biases
\end{enumerate}

These criteria act as useful guides for working towards a benchmark for dialogue on robots. We add here several requirements for interactive dialogue with robots that a benchmark should fulfill:

    \paragraph{Multimodal} the data needs to have speech (or text), but also information about the physical situation including sensor information from sensors such as one or more cameras, as well as robot state information.
    \paragraph{Co-located} the data needs to contain dialogue that talks about objects in a shared environment that the robot can manipulate (e.g., identify, grasp, or move).
    \paragraph{High-stakes Dialogue} the task needs to require that the user and robot work together to accomplish something in a scenario where the stakes are sufficiently high so as to require important dialogue artifacts such as clarification requests, interruptions, building common ground, and providing explanations of decisions. For example, a fast-paced or gamified scenario increases stakes above simple, goalless verbal interaction. 
    \paragraph{User-centered} the user needs to feel like they are not just giving the robot commands; rather, they see the benefit for both the user and the robot's involvement. 
    \paragraph{Community-agnostic} the benchmark should use different robot platforms, be affordable, work in both a virtual (including the possibility of VR headset integrations) and real environments. 

Other properties of the benchmark could include the ability to evaluate scenarios where multiple robots and/or people are participating, learning during interaction, and focus is put on underrepresented communities.  

\subsection{Existing Work}

\begin{figure*}
\centering
\includegraphics[width=0.7\textwidth]{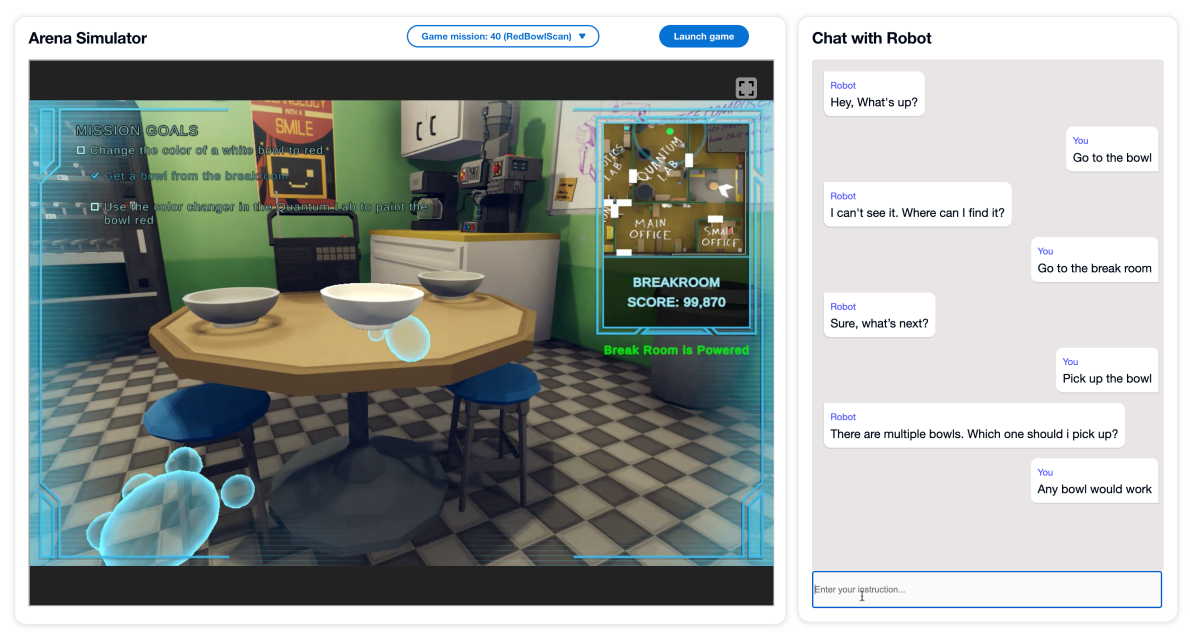}
\caption{\label{fig:alexarena} Example of the Alexa Arena benchmark: user utterances and simulated robot responses, from \citet{Gao2023-ed}}
\end{figure*}

Benchmarks are not new in \sds\ research. The Dialogue State Tracking Challenge, for example, ran for several years \cite{Williams2016-nj,Kim2017-rz}. There were multiple tasks over the years including finding bus timetable information and restaurant search. However, the goal of dialogue state tracking is only part of what makes a dialogue system work, and the only modality was transcribed speech (i.e., text).

Closer to meeting the above requirements is the ALFRED (Action Learning from Realistic Environments and Directives) benchmark for learning a mapping from natural language instructions and egocentric vision to sequences of actions for household tasks such as navigating and moving objects \cite{Shridhar2019-sz}. The world is represented virtually, but the agent can freely move around in a complex environment. Similar in some ways to ALFRED, the TEACh benchmark \cite{Padmakumar2021-jx} uses a \emph{commander} and \emph{follower} paradigm where human users take the commander role and instruct the robot follower in a household setting performing tasks of varying complexity. In both challenges, the primary interaction is through a text-input chat interface, clarification requests are part of the TEACh dataset. 

Similar but closer to the above requirements is Alexa Arena \cite{Gao2023-ed}, see Figure~\ref{fig:alexarena}. The platform is a new user-centric ``Embodied" AI platform which focuses on building generalizable agents through reasoning and procedural learning, a benchmark of 3k unique tasks (e.g., move, lift) and 46k human-annotated instructions and dialogues. The platform includes a web-based user interface for easy integration and real-time user interaction (see Figure~\ref{fig:alexarena}). Simple ``verbal joystick” actions are possible such as \emph{turn left} or \emph{move forward}, but more complex, relational referring expressions and interactive dialogue are also required to complete tasks. This dataset is interesting to researchers because there are many breakdowns in communication that require \sds, the setting is complex enough to require complex language, yet simple enough to not be completely open domain. Robots are not required as the setting is virtual. 

These benchmarks offer a valuable starting point for a dialogue-with-robots benchmark, but they do not fulfill all of the requirements listed above. Below we sketch a proposal towards the goal of a benchmark that does. 

\subsection{Proposal}

The work in developing the benchmark will have three phases: (1) requirements gathering (what do researchers need in a benchmark?), (2) developing technical infrastructure including virtual platforms, and  (3) establishing the first cohort of participants in a challenge that uses the benchmark. 

\paragraph{Requirements Gathering} Research is driven by research questions. The kinds of robots and tasks that are interesting to particular researchers dictates the kinds of robots and platforms that they adopt. Some are more focused on robot navigation, others on language/symbol grounding, still others put more focus on the social aspects of interaction. In any case, the benchmark should include aspects of safety and bias. 

\begin{figure*}[t]
    \centering
    \includegraphics[width=.7\textwidth]{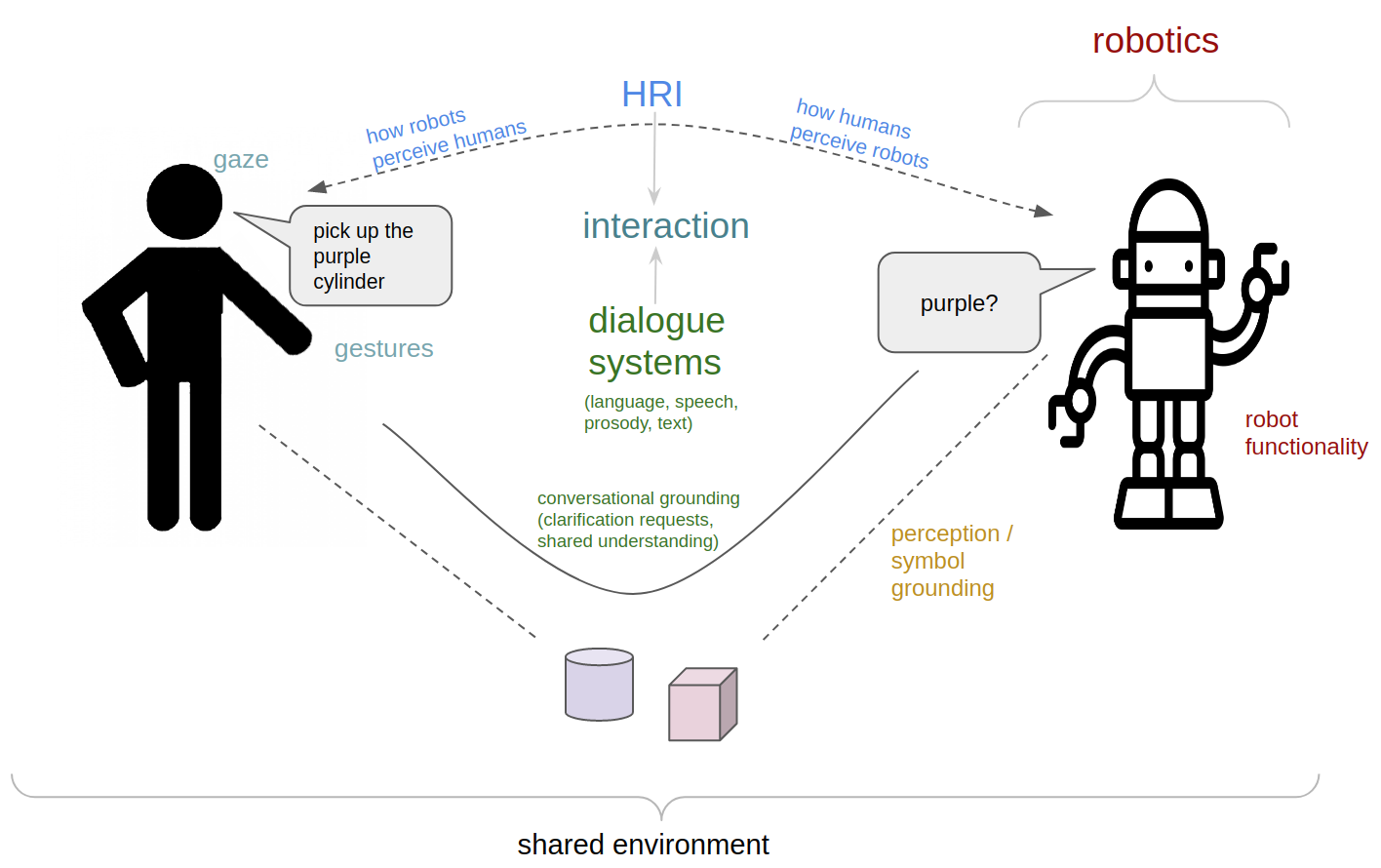}
    \caption{The fields of robotics, \hri, and \sds\ have different, yet complementary goals, from \citet{Kennington2021-qh}.}
    \label{fig:goals}
\end{figure*}

The proposed work will involve requirements gathering to determine what platforms and tasks researchers are using, and what kind of research they are focused on. \citet{Kennington2021-qh} noted that the fields of robotics, \sds, and \hri\ have different, yet complementary goals (see Figure~\ref{fig:goals}) to enable robots and humans to work together more naturally and effectively, which offers a general starting point. Though requirements for each researcher will be different, the design of the infrastructure can have some degree of navigation, dialogue interaction including social aspects, and interaction with objects in the environment for language/symbol grounding (i.e., connecting language to the physical world).

\paragraph{Building Infrastructure} The challenge should include a virtual (e.g., similar to Alexa Arena) and real-world version of the the same task. The virtual version will enable more researchers access to the benchmark without expensive hardware. The real-world version will serve as proving grounds that the benchmark will work in a real robot. 

Similar in some ways to DuckieTown which offers a robot platform and environment (though DuckieTown is focused on navigation), the platform should make the environment available to the community.\footnote{\url{https://www.duckietown.org}} Our goal is to allow as many objects to be easily obtained (e.g., through 3D printing) at low cost and to allow any robot platform within certain size dimensions, and we will maintain a public listing of robots being used for the challenge. 

\paragraph{Challenge} We suggest that there be an initial challenge that helps guide a cohort of teams through the benchmark. This will serve as a test for the benchmark as the teams will receive materials without cost and they will be able to help work through inevitable technical difficulties. This approach follows recent NSF-backed research infrastructure to build a modular social robot.\footnote{\url{https://www.nsf.gov/awardsearch/showAward?AWD_ID=2235042&HistoricalAwards=false}}

The first challenge should begin with a cohort of 5 teams to beta-test the challenge with real researchers. The teams will be chosen based on their experience in the field of \sds\ and \hri\ and their proposal of the method they will use to fulfill the challenge. Each team will be issued the virtual environment, a robot (yet to be determined; we look to GoPiGo as a possible starting point because it is cheap and simple, yet can sense the environment and move around in it), and the environment. There will be a setup, training, and evaluation phase of the first challenge. Teams will be able to report the results of their work in a culminating workshop that will serve as a report for the first cohort and an open invitation for others interested in using the benchmark.

\section{Proposal 3: Large Language Models and Robots}


\emph{Large Language Models} (\llm) have been a mainstay of \nlp\ since attention neural mechanisms used in a transformer \cite{Vaswani2017-kv} and the BERT transformer were introduced and systematically used for language tasks \cite{Devlin2018-lz}. \llm s like BERT use a pre-train, fine-tune paradigm: during pre-training, the training regime tasks the model with guessing \emph{masked} words within a sequence of words (i.e., they can be in a lexical context or a next word/sentence to be predicted). This training regime enables the model to learn a distributional approximation of meanings of words and sentences from the text they are trained on. During the fine-tuning phase, the models can be arranged in such a way that they can take in text as input and be tuned to a specific task like machine translation, sentiment classification, or topic modeling. \llm s effectively allow one to train using a straightforward training regime where only text is required, then tune to specific tasks where only little data might be available. 

Can \llm s be used for dialogue with robots? A social robot could straight-forwardly be hooked up to a speech recognizer which transcribes speech, then that transcription is given to a language model which produces a response. This is fine for social tasks where the goal is chatting, but the limits of training only on text, and the ability for \llm s to only take text as input are quickly apparent when one wants the robot to refer to objects in a physical environment---\llm s trained only on text have no notion of the physical world unless the physical world is somehow represented in text symbols as input to the \llm. Being trained on text only also hasn't prevented \llm s from potential ethical issues such as generating abusive, stereotyped or biased language, or fabricating information that is nonfactual \cite{Bender2021-gg}. 
Moreover, the SLIVAR workshop on the Ethics of Language-Capable Robots identified a range of key ethical challenges facing roboticists developing language-capable robots, especially those hoping to use LLMs as part of their robotics solutions~\cite{williams2023voice}. These include key challenges surrounding trust and influence in and by language capable-robots; identity performance by language-capable robots; and privacy concerns surrounding what is heard and/or understood by language-capable robots.
Mitigating these shortcomings in \llm s is an ongoing research trend by the broader community~\cite{williams2024scarecrows}, but it is something that will need to be taken into account and addressed in the setting of human-robot dialogue.

A more recent trend in \llm\ advancement has been to enrich language models with visual information, known as \emph{multi-modal language models} (\mmllm). For example, the VilBERT model uses BERT coupled with a visual representation of images \cite{Lu2019-ex}. Others have proposed language and vision tasks to evaluate such models, such as Visual Dialogue \cite{Das2019-sh} and Visual Question Answering \cite{Antol2015-fs}. More recent work has explored how different training regimes and architectures work better for vision and language tasks \cite{Dou2022-uo}. More recent ``embodied'' models allow for incorporating modalities beyond text (and vision) into the language model, but are yet to be tested within human-robot interaction studies \cite{Driess2023-sf}. More directly for robots, \cite{Gao2023-pd} showed how vision-language \llm s can be used for robotic manipulation of objects. Finally, the One-Peace model \citet{Wang2023-fa} can fold any input modality into the self-attention steps of a \llm, tested on vision and audio, but theoretically could be used with other modalities important to robots such as internal state representations. The research cited here represents steps in the right direction because they open the visual world up to \llm s which is crucial for robots. 

Beyond spoken interaction or visual language tasks, \llm s are being used for any sequential task or mapping from language to some kind of logical form (which robots are often represented by). For example, \cite{Schick2023-jc} introduced the \emph{Toolformer} model that can map from language to `tool` access (i.e., API calls) for example, if numbers are detected in a sentence, the Toolformer would call a calculator tool, or if a word in a foreign language is detected then the Toolformer calls a translation tool. This is useful for robotic platforms where if someone utters an instruction (e.g., \emph{pick up the book}), a \llm\ like the Toolformer could map that to the tool (e.g., arm movement) that is necessary to complete the task. 

\citet{Zhang2023-ko} used language models to reason about a situation in order to build cooperative embodied (in their case, virtual) agents. For example, one agent can inform another agent about objects in a room and the other agent can take actions on those objects. The inform step is natural language generated by a language model. \citet{Min2021-ms} introduced \emph{FILM} that maps language sequences to action sequence subtasks (e.g., \emph{(Pan, PickUp)}) that is then combined with visual input, then the agent can actually take actions. Another model, ProgPrompt, can map from a natural language prompt to python-like programming language instructions that can then be directly executed to control a robot \cite{Singh2023-ea}. Language models have even been used to generation emotional behaviors on a robot platform \cite{Mishra2023-wp}. 

Taken together, there are two ways that researchers make use of language models:
\begin{itemize}
    \item Convert the physical world into tokens, then fine-tune a language model on the input/output at the level of symbolic tokens (e.g., FILM and ProgPrompt)
    \item Represent the physical world a fine-grained level (usually in vector form) and fuse within the model itself (e.g., One-Peace) 
\end{itemize}

\llm s are clearly a model of choice for a growing number of tasks for robots ranging from observing the visual world, interacting with human users, and generating emotional displays. \llm s are well suited for sequential tasks. However, \llm s do need a full input sequence to map to a desired output label or sequence. \llm s do not work \emph{incrementally} in that they cannot simply accept a single token at a time and produce output tokens monotonically. As a result, they are not inherently temporal in nature, which is important in real-time interaction. However, see recent work in making transformer language models work incrementally (i.e., word-by-word) \cite{Kahardipraja2021-fv}. 

\subsection{Gaps}

\paragraph{Closed LLMs} \llm s like ChatGPT and GPT-4 are impressive and useful, but it is not obvious how to hook them up to live robots without encoding everything about an ongoing interaction between a human and robot symbolically using text and coming up with a textual response that is not only what the robot should say, but also what the robot should do. Additionally, they have security and bias issues. Moreover, the latency for getting a response from a public \llm\ is usually too long for co-located interaction to be effective. Finally, as \citet{Rogers2023-up} points out, closed \llm s (\emph{closed} meaning we don't know what the architectures are or what data they were trained on) make bad baselines to compare against for ongoing research because much about them are not sufficiently explained as to be reproducible. 

\paragraph{Model Size} \llm s are called \emph{large} for a reason. BERT has 110 million parameters \cite{Devlin2018-lz} and RoBERTa has 125 million parameters \cite{Liu2019-so}. These are more manageable than, say, ChatGPT (estimated around 175 billion parameters), but the compute resources required for pre-training can be out of reach to many researchers. One promising model for research involving robots and language is Palm-E, a model that incorporates multiple modalities beyond just text, and which is based on the PALM model \cite{Chowdhery2022-qs} required over 6,000 TPUv3 chips (a cost of roughly \$20,000 per hour) to train. These kinds of models that require excessive compute resources limit the kinds of fundamental research that is needed for human-robot dialogue. While fine-tuning such models on specific tasks is a common practice, there are many open research questions in the domain of human-robot dialogue that will require custom pre-training, but work needs to be done to make the models small, yet effective. 

\paragraph{Data Size} The amount of data needed to train \mmllm s is, as the number of model parameters, a lot. Text data, and even vision data which are currently used to train \mmllm s are much easier to come by than data recording robot behaviors and states, as well as interactions between robots and humans. The kinds of systems required are complex, making data collection challenging, resulting in very little effective data. \llm s tend to be general purpose models, but what is needed in dialogue with robots is specific purpose: robots need to interact with humans using speech, but also attend to objects using other robot modalities such as pointing or grasping. Without effective learning on small amounts of data, it will be difficult to advance important research in this area. 

Relatedly, many \llm s are trained end-to-end with clear text input, text output tasks. For example, for question-answering there is an input question and an output answer. For dialogue, there is an input utterance and an output response. Can end-to-end modeling be used for human-robot interaction where dialogue is present? It is possible, but data is required---a lot of data. As noted above, it is also not clear how to ``hook up" the modalities of the robot (cameras, bump sensors, depth sensors, internal states) and the ongoing interaction to the end-to-end \mmllm. At the moment, it is more prudent to construct systems that are modular where \llm s are present, perhaps to understand utterances and generate responses (explored below). 

\subsection{Open Questions} 

To address the above gaps, we propose research on small language models that require fewer compute resources to train and run in real-time, but also require less data to be effective. This goal is of growing concern in the broader community given new smaller language models like BabyBERTa \cite{Huebner2021-ri}, distillation \cite{kim2022soda,Hsieh2023-kf}, and a recent challenge for BabyLMs that require less training data \cite{Warstadt2023-gv}. We echo the need for these efforts and encourage research in this direction. 

More work needs to be done to see how \llm s, particularly \mmllm s can work together with robots in a dialogue setting. \llm s are already being used for language understanding tasks in dialogue interaction with robots, and they are already being used to generate responses. 

\begin{figure*}
    \centering
    \includegraphics[width=0.5\paperwidth]{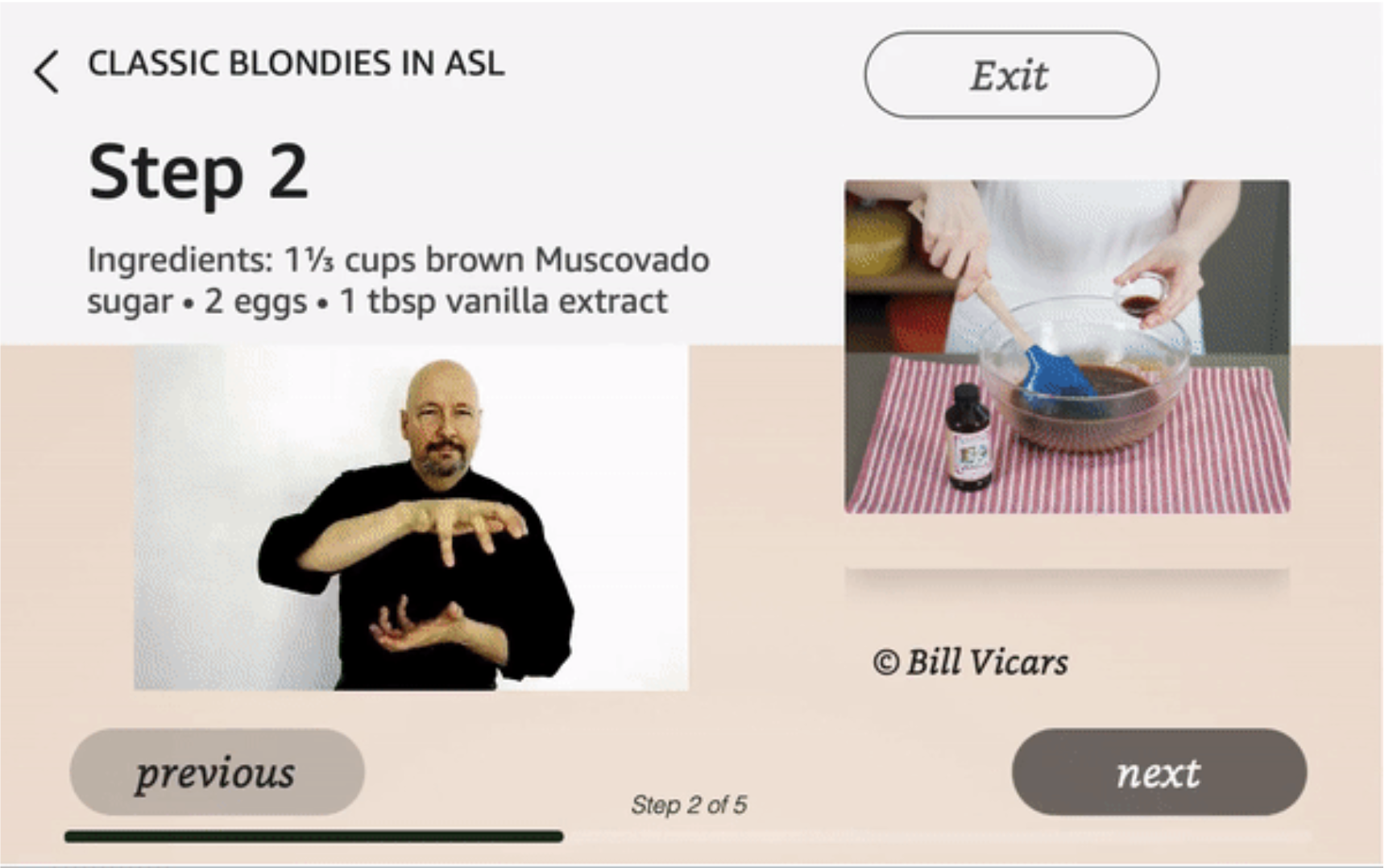}
    \caption{Example of the ISABEL bot designed as part of the Amazon Alexa Taskbot Challenge 2023. ISABEL is the first multimodal bot that can interact with signers around the world \textcolor{red}{add the role of LLM Malihe} \cite{Pittsburgh2023}. }
    \label{fig:signalexa}
\end{figure*}
One open question is how the world and unfolding interaction can (or should) be encoded for the robot to effectively interact. \llm s require text as input. Should models of the world be encoded in symbolic text, and if so, how? What is the ``vocabulary" of representing the world that would be effective for \llm s? Such an approach could be effective because it would not require costly remodelling of the \llm\ to include new modalities, though researchers would need to find new ways to textually represent newly added modalities (e.g., a bump sensor). \citet{Williams2024-ng} called \llm s that are used as stand-ins for modules 'scarecrows' in that they are used as black-box modules to enable full-pipeline solutions. For example, using a \llm\ for modeling what a human might say to the robot is a scarecrow. These kinds of uses of \llm s have some theoretical implications, for example that using a text-only \llm\ to make decisions about all aspects of an embodied interaction makes a tacit assumption that all knowledge and decisions are language-based (e.g., following the Sapir-Whorf hypothesis). Moreover, and more importantly, the issues with bias that exist in pre-trained \llm s would negatively affect the qualities of the interaction, particularly when the humans involved are part of underrepresented groups, which are often the focus of human-robot interaction research (e.g., autistic children or teenagers with mental illnesses). 

Alternatively, similar to the Palm-E and OnePeace models, information from each modality can be embedded in a vector and incorporated into the model. This has some theoretical implications in that it encodes knowledge and decisions not only as text. It also makes modeling more difficult as new modality need to somehow be represented vectorally and incorporated into the model, and the model needs enough data to learn how to make use of the information from the new modality, negating the effectiveness of some pre-traind models. 

One promising line of work that will affect human-robot dialogue is recent work on \emph{reinforcement learning with human feedback} (\rlhf). Reinforcement learning in \sds\ research is not new, spanning back to early efforts in dialogue state tracking \cite{schlangen2003coherence}, and effective learning from humans while interacting has long been a goal of \sds\ and \hri\ research, such as learning from explicit examples. The explicit learning approaches used in \rlhf\ have direct implications for human-robot dialogue as robots are often tasked with learning as they verbally interact with humans. 

\subsection{Proposal}
We encourage work on language model research that (1) focuses on smaller, yet effective LLMs (particularly MLLMs) that can work on a small piece of hardware on a robot \cite{sicilia2022leather}, (2) focuses on effective learning with small amounts of data, (3) focuses on mitigating the biases found in LLMs (e.g., \citet{dana2023socially,sicilia2023learning}), (4) support populations with diverse needs and capabilities (e.g., \citet{yin2021including,viegas2022including,inan2022modeling,tong2024eyes}), and (5) support users in achieving their goals and common ground with them.

Concerning bias, safety, and inclusivity, it is paramount to address these aspects thoroughly within the context of LLM development. Bias mitigation strategies are essential to counteract and reduce the negative impacts of inherent biases present in LLMs, necessitating recent research that identifies and corrects these biases effectively \cite{dana2023socially,weidinger2023sociotechnical} this includes grounding our models in theoretical foundations of machine learning that can help us learn from the long tail \cite{sicilia2023learning}, active learning approaches for data augmentation and training \cite{hassan2023d} and employing cognitively motivated interventions such as generating paraphrases and counter narratives \cite{atwell2022appdia,hassan2023discgen}.  Safety measures must be integrated to ensure that LLMs do not produce or propagate harmful, misleading, or inapproprmoreiate content, thereby protecting users from potential risks associated with automated language generation. Inclusivity requires the design and development of LLMs that accommodate and support the varied needs and abilities of diverse user populations \cite{yin2021including,viegas2022including,inan2022modeling,tong2024eyes}, ensuring accessibility and utility across different socio-cultural and linguistic contexts. Figure\ref{fig:signalexa} shows one of the few multimodal dialogue that can interact with signers. This is just one example of ways we can be thinking about accommodating needs of our diverse community of users. This holistic approach towards bias, safety, and inclusivity underscores the importance of creating LLMs that are not only technologically sophisticated but also ethically sound and socially responsible.

\section{Conclusion}

This paper synthesizes the discussions from the NSF-funded workshop in 2023 on the topic of robots and dialogue. The three proposals above represent the ongoing and future work that we urge the community to address in the near future. 

As coursework evolves to train new graduates that are ready for \sds-\hri\ research, more fundamental research and practical applications will enable people to be more productive and ready to interact with robots in the workforce. As benchmarks for spoken dialogue with robots improve, models and datasets will follow that build on those improvements. As large language models become more multi-modal, process in real time, and learn from more than just text as they interact, their usefulness as a model of language understanding and generation with robots will increase. 

We are optimistic that the field of spoken interaction with robots will continue to grow and impact society in positive ways. 

\paragraph{Acknoledgements} This workshop was funded by the National Science Foundation Award \#2306113. We want to thank those who participated in other ways, including Erin Walker (University of Pittsburgh), Dan Bohus (Microsoft Research), and Joyce Chai (University of Michigan). We want to thank the University of Pittsburgh for hosting the event, and thanks to the administrative staff who helped organized the workshop.

\bibliography{custom,paperpile}

\begin{thebibliography}{50}
\expandafter\ifx\csname natexlab\endcsname\relax\def\natexlab#1{#1}\fi

\bibitem[{Antol et~al.(2015)Antol, Agrawal, Lu, Mitchell, Batra, Zitnick, and
  Parikh}]{Antol2015-fs}
Stanislaw Antol, Aishwarya Agrawal, Jiasen Lu, Margaret Mitchell, Dhruv Batra,
  C~Lawrence Zitnick, and Devi Parikh. 2015.
\newblock {VQA}: Visual question answering.
\newblock In \emph{Proceedings of the {IEEE} International Conference on
  Computer Vision}.

\bibitem[{Atwell et~al.(2022)Atwell, Hassan, and Alikhani}]{atwell2022appdia}
Katherine Atwell, Sabit Hassan, and Malihe Alikhani. 2022.
\newblock Appdia: A discourse-aware transformer-based style transfer model for
  offensive social media conversations.
\newblock \emph{arXiv preprint arXiv:2209.08207}.

\bibitem[{Bender et~al.(2021)Bender, Gebru, McMillan-Major, and
  Shmitchell}]{Bender2021-gg}
Emily~M Bender, Timnit Gebru, Angelina McMillan-Major, and Shmargaret
  Shmitchell. 2021.
\newblock On the dangers of stochastic parrots: Can language models be too big?
\newblock In \emph{Proceedings of the 2021 {ACM} Conference on Fairness,
  Accountability, and Transparency}, FAccT '21, pages 610--623, New York, NY,
  USA. Association for Computing Machinery.

\bibitem[{Bowman and Dahl(2021)}]{Bowman2021-xn}
Samuel~R Bowman and George Dahl. 2021.
\newblock What will it take to fix benchmarking in natural language
  understanding?
\newblock In \emph{Proceedings of the 2021 Conference of the North American
  Chapter of the Association for Computational Linguistics: Human Language
  Technologies}, pages 4843--4855, Online. Association for Computational
  Linguistics.

\bibitem[{Chowdhery et~al.(2022)Chowdhery, Narang, Devlin, Bosma, Mishra,
  Roberts, Barham, Chung, Sutton, Gehrmann, Schuh, Shi, Tsvyashchenko, Maynez,
  Rao, Barnes, Tay, Shazeer, Prabhakaran, Reif, Du, Hutchinson, Pope, Bradbury,
  Austin, Isard, Gur-Ari, Yin, Duke, Levskaya, Ghemawat, Dev, Michalewski,
  Garcia, Misra, Robinson, Fedus, Zhou, Ippolito, Luan, Lim, Zoph, Spiridonov,
  Sepassi, Dohan, Agrawal, Omernick, Dai, Pillai, Pellat, Lewkowycz, Moreira,
  Child, Polozov, Lee, Zhou, Wang, Saeta, Diaz, Firat, Catasta, Wei,
  Meier-Hellstern, Eck, Dean, Petrov, and Fiedel}]{Chowdhery2022-qs}
Aakanksha Chowdhery, Sharan Narang, Jacob Devlin, Maarten Bosma, Gaurav Mishra,
  Adam Roberts, Paul Barham, Hyung~Won Chung, Charles Sutton, Sebastian
  Gehrmann, Parker Schuh, Kensen Shi, Sasha Tsvyashchenko, Joshua Maynez,
  Abhishek Rao, Parker Barnes, Yi~Tay, Noam Shazeer, Vinodkumar Prabhakaran,
  Emily Reif, Nan Du, Ben Hutchinson, Reiner Pope, James Bradbury, Jacob
  Austin, Michael Isard, Guy Gur-Ari, Pengcheng Yin, Toju Duke, Anselm
  Levskaya, Sanjay Ghemawat, Sunipa Dev, Henryk Michalewski, Xavier Garcia,
  Vedant Misra, Kevin Robinson, Liam Fedus, Denny Zhou, Daphne Ippolito, David
  Luan, Hyeontaek Lim, Barret Zoph, Alexander Spiridonov, Ryan Sepassi, David
  Dohan, Shivani Agrawal, Mark Omernick, Andrew~M Dai,
  Thanumalayan~Sankaranarayana Pillai, Marie Pellat, Aitor Lewkowycz, Erica
  Moreira, Rewon Child, Oleksandr Polozov, Katherine Lee, Zongwei Zhou, Xuezhi
  Wang, Brennan Saeta, Mark Diaz, Orhan Firat, Michele Catasta, Jason Wei,
  Kathy Meier-Hellstern, Douglas Eck, Jeff Dean, Slav Petrov, and Noah Fiedel.
  2022.
\newblock \href {http://arxiv.org/abs/2204.02311} {{PaLM}: Scaling language
  modeling with pathways}.

\bibitem[{Dana et~al.(2023)Dana, Andrews, Bekris, Feldman, Stone, Hemmer,
  Mazzeo, Salzman, and Yi}]{dana2023socially}
Kristin~J Dana, Clinton Andrews, Kostas Bekris, Jacob Feldman, Matthew Stone,
  Pernille Hemmer, Aaron Mazzeo, Hal Salzman, and Jingang Yi. 2023.
\newblock Socially cognizant robotics for a technology enhanced society.
\newblock \emph{arXiv preprint arXiv:2310.18303}.

\bibitem[{Das et~al.(2019)Das, Kottur, Gupta, Singh, Yadav, Lee, Moura, Parikh,
  and Batra}]{Das2019-sh}
Abhishek Das, Satwik Kottur, Khushi Gupta, Avi Singh, Deshraj Yadav, Stefan
  Lee, Jose M~F Moura, Devi Parikh, and Dhruv Batra. 2019.
\newblock Visual dialog.
\newblock \emph{IEEE Trans. Pattern Anal. Mach. Intell.}, 41(5).

\bibitem[{Devlin et~al.(2018)Devlin, Chang, Lee, and Toutanova}]{Devlin2018-lz}
Jacob Devlin, Ming-Wei Chang, Kenton Lee, and Kristina Toutanova. 2018.
\newblock {BERT}: Pre-training of deep bidirectional transformers for language
  understanding.

\bibitem[{Dou et~al.(2022)Dou, Xu, Gan, Wang, Wang, Wang, Zhu, Zhang, Yuan,
  Peng, Liu, and Zeng}]{Dou2022-uo}
Zi-Yi Dou, Yichong Xu, Zhe Gan, Jianfeng Wang, Shuohang Wang, Lijuan Wang,
  Chenguang Zhu, Pengchuan Zhang, Lu~Yuan, Nanyun Peng, Zicheng Liu, and
  Michael Zeng. 2022.
\newblock An empirical study of training end-to-end vision-and-language
  transformers.
\newblock In \emph{2022 {IEEE/CVF} Conference on Computer Vision and Pattern
  Recognition ({CVPR})}, pages 18166--18176. IEEE.

\bibitem[{Driess et~al.(2023)Driess, Xia, Sajjadi, Lynch, Chowdhery, Ichter,
  Wahid, Tompson, Vuong, Yu, Huang, Chebotar, Sermanet, Duckworth, Levine,
  Vanhoucke, Hausman, Toussaint, Greff, Zeng, Mordatch, and
  Florence}]{Driess2023-sf}
Danny Driess, Fei Xia, Mehdi S~M Sajjadi, Corey Lynch, Aakanksha Chowdhery,
  Brian Ichter, Ayzaan Wahid, Jonathan Tompson, Quan Vuong, Tianhe Yu, Wenlong
  Huang, Yevgen Chebotar, Pierre Sermanet, Daniel Duckworth, Sergey Levine,
  Vincent Vanhoucke, Karol Hausman, Marc Toussaint, Klaus Greff, Andy Zeng,
  Igor Mordatch, and Pete Florence. 2023.
\newblock \href {http://arxiv.org/abs/2303.03378} {{PaLM-E}: An embodied
  multimodal language model}.

\bibitem[{Eskenazi and Zhao(2020)}]{Eskenazi2020-no}
Maxine Eskenazi and Tiancheng Zhao. 2020.
\newblock Report from the {NSF} future directions workshop, toward
  {User-Oriented} agents: Research directions and challenges.
\newblock \emph{arXiv}.

\bibitem[{Gao et~al.(2023{\natexlab{a}})Gao, Sarkar, Xia, Xiao, Wu, Ichter,
  Majumdar, and Sadigh}]{Gao2023-pd}
Jensen Gao, Bidipta Sarkar, Fei Xia, Ted Xiao, Jiajun Wu, Brian Ichter,
  Anirudha Majumdar, and Dorsa Sadigh. 2023{\natexlab{a}}.
\newblock \href {http://arxiv.org/abs/2309.02561} {Physically grounded
  {Vision-Language} models for robotic manipulation}.

\bibitem[{Gao et~al.(2023{\natexlab{b}})Gao, Thattai, Gao, Shakiah, Pansare,
  Sharma, Sukhatme, Shi, Yang, Zheng, Hu, Arumugam, Hu, Wen, Guthy, Chung,
  Khanna, Ipek, Ball, Bland, Rocker, Rao, Johnston, Ghanadan, Mandal, Tur, and
  Natarajan}]{Gao2023-ed}
Qiaozi Gao, Govind Thattai, Xiaofeng Gao, Suhaila Shakiah, Shreyas Pansare,
  Vasu Sharma, Gaurav Sukhatme, Hangjie Shi, Bofei Yang, Desheng Zheng, Lucy
  Hu, Karthika Arumugam, Shui Hu, Matthew Wen, Dinakar Guthy, Cadence Chung,
  Rohan Khanna, Osman Ipek, Leslie Ball, Kate Bland, Heather Rocker,
  Yadunandana Rao, Michael Johnston, Reza Ghanadan, Arindam Mandal,
  Dilek~Hakkani Tur, and Prem Natarajan. 2023{\natexlab{b}}.
\newblock \href {http://arxiv.org/abs/2303.01586} {Alexa arena: A
  {User-Centric} interactive platform for embodied {AI}}.

\bibitem[{Hassan and Alikhani(2023{\natexlab{a}})}]{hassan2023d}
Sabit Hassan and Malihe Alikhani. 2023{\natexlab{a}}.
\newblock D-calm: A dynamic clustering-based active learning approach for
  mitigating bias.
\newblock \emph{arXiv e-prints}, pages arXiv--2305.

\bibitem[{Hassan and Alikhani(2023{\natexlab{b}})}]{hassan2023discgen}
Sabit Hassan and Malihe Alikhani. 2023{\natexlab{b}}.
\newblock \href {http://arxiv.org/abs/2311.18147} {Discgen: A framework for
  discourse-informed counterspeech generation}.

\bibitem[{Hsieh et~al.(2023)Hsieh, Li, Yeh, Nakhost, Fujii, Ratner, Krishna,
  Lee, and Pfister}]{Hsieh2023-kf}
Cheng-Yu Hsieh, Chun-Liang Li, Chih-Kuan Yeh, Hootan Nakhost, Yasuhisa Fujii,
  Alexander Ratner, Ranjay Krishna, Chen-Yu Lee, and Tomas Pfister. 2023.
\newblock \href {http://arxiv.org/abs/2305.02301} {Distilling {Step-by-Step}!
  outperforming larger language models with less training data and smaller
  model sizes}.

\bibitem[{Huebner et~al.(2021)Huebner, Sulem, Cynthia, and
  Roth}]{Huebner2021-ri}
Philip~A Huebner, Elior Sulem, Fisher Cynthia, and Dan Roth. 2021.
\newblock {{B}aby{BERT}a}: Learning more grammar with {Small-Scale}
  {Child-Directed} language.
\newblock In \emph{Proceedings of the 25th Conference on Computational Natural
  Language Learning}, pages 624--646, Online. Association for Computational
  Linguistics.

\bibitem[{Inan et~al.(2022)Inan, Zhong, Hassan, Quandt, and
  Alikhani}]{inan2022modeling}
Mert Inan, Yang Zhong, Sabit Hassan, Lorna Quandt, and Malihe Alikhani. 2022.
\newblock Modeling intensification for sign language generation: A
  computational approach.
\newblock In \emph{Findings of the Association for Computational Linguistics:
  ACL 2022}, pages 2897--2911.

\bibitem[{Kahardipraja et~al.(2021)Kahardipraja, Madureira, and
  Schlangen}]{Kahardipraja2021-fv}
Patrick Kahardipraja, Brielen Madureira, and David Schlangen. 2021.
\newblock Towards incremental transformers: An empirical analysis of
  transformer models for incremental {NLU}.
\newblock In \emph{Proceedings of the 2021 Conference on Empirical Methods in
  Natural Language Processing}, pages 1178--1189, Online and Punta Cana,
  Dominican Republic. Association for Computational Linguistics.

\bibitem[{Kennington(2021{\natexlab{a}})}]{Kennington2021-uj}
Casey Kennington. 2021{\natexlab{a}}.
\newblock Natural language processing for computer scientists and data
  scientists at a large state university.
\newblock In \emph{Proceedings of the Fifth Workshop on Teaching {NLP}}, pages
  115--124, Online. Association for Computational Linguistics.

\bibitem[{Kennington(2021{\natexlab{b}})}]{Kennington2021-qh}
Casey Kennington. 2021{\natexlab{b}}.
\newblock \href {http://arxiv.org/abs/2108.10931} {The state of {SLIVAR}:
  What's next for robots, human-robot interaction, and (spoken) dialogue
  systems?}

\bibitem[{Kim et~al.(2022)Kim, Hessel, Jiang, West, Lu, Yu, Zhou, Bras,
  Alikhani, Kim et~al.}]{kim2022soda}
Hyunwoo Kim, Jack Hessel, Liwei Jiang, Peter West, Ximing Lu, Youngjae Yu, Pei
  Zhou, Ronan~Le Bras, Malihe Alikhani, Gunhee Kim, et~al. 2022.
\newblock Soda: Million-scale dialogue distillation with social commonsense
  contextualization.
\newblock \emph{arXiv preprint arXiv:2212.10465}.

\bibitem[{Kim et~al.(2017)Kim, D'Haro, Banchs, Williams, and
  Henderson}]{Kim2017-rz}
Seokhwan Kim, Luis~Fernando D'Haro, Rafael~E Banchs, Jason~D Williams, and
  Matthew Henderson. 2017.
\newblock The fourth dialog state tracking challenge.
\newblock In \emph{Dialogues with Social Robots}, pages 435--449. Springer.

\bibitem[{Liu et~al.(2019)Liu, Ott, Goyal, Du, Joshi, Chen, Levy, Lewis,
  Zettlemoyer, and Stoyanov}]{Liu2019-so}
Yinhan Liu, Myle Ott, Naman Goyal, Jingfei Du, Mandar Joshi, Danqi Chen, Omer
  Levy, Mike Lewis, Luke Zettlemoyer, and Veselin Stoyanov. 2019.
\newblock {RoBERTa}: A robustly optimized {BERT} pretraining approach.

\bibitem[{Lu et~al.(2019)Lu, Batra, Parikh, and Lee}]{Lu2019-ex}
Jiasen Lu, Dhruv Batra, Devi Parikh, and Stefan Lee. 2019.
\newblock {ViLBERT}: Pretraining {Task-Agnostic} visiolinguistic
  representations for {Vision-and-Language} tasks.

\bibitem[{Marge et~al.(2022)Marge, Espy-Wilson, Ward, Alwan, Artzi, Bansal,
  Blankenship, Chai, Daum{\'e}, Dey, Harper, Howard, Kennington,
  Kruijff-Korbayov{\'a}, Manocha, Matuszek, Mead, Mooney, Moore, Ostendorf,
  Pon-Barry, Rudnicky, Scheutz, Amant, Sun, Tellex, Traum, and
  Yu}]{Marge2022-yf}
Matthew Marge, Carol Espy-Wilson, Nigel~G Ward, Abeer Alwan, Yoav Artzi, Mohit
  Bansal, Gil Blankenship, Joyce Chai, Hal Daum{\'e}, Debadeepta Dey, Mary
  Harper, Thomas Howard, Casey Kennington, Ivana Kruijff-Korbayov{\'a}, Dinesh
  Manocha, Cynthia Matuszek, Ross Mead, Raymond Mooney, Roger~K Moore, Mari
  Ostendorf, Heather Pon-Barry, Alexander~I Rudnicky, Matthias Scheutz,
  Robert~St Amant, Tong Sun, Stefanie Tellex, David Traum, and Zhou Yu. 2022.
\newblock Spoken language interaction with robots: Recommendations for future
  research.
\newblock \emph{Comput. Speech Lang.}, 71:101255.

\bibitem[{Min et~al.(2021)Min, Chaplot, Ravikumar, Bisk, and
  Salakhutdinov}]{Min2021-ms}
So~Yeon Min, Devendra~Singh Chaplot, Pradeep Ravikumar, Yonatan Bisk, and
  Ruslan Salakhutdinov. 2021.
\newblock \href {http://arxiv.org/abs/2110.07342} {{FILM}: Following
  instructions in language with modular methods}.

\bibitem[{Mishra et~al.(2023)Mishra, Verdonschot, Hagoort, and
  Skantze}]{Mishra2023-wp}
Chinmaya Mishra, Rinus Verdonschot, Peter Hagoort, and Gabriel Skantze. 2023.
\newblock Real-time emotion generation in human-robot dialogue using large
  language models.
\newblock \emph{Front Robot AI}, 10:1271610.

\bibitem[{Padmakumar et~al.(2021)Padmakumar, Thomason, Shrivastava, Lange,
  Narayan-Chen, Gella, Piramithu, Tur, and Hakkani-Tur}]{Padmakumar2021-jx}
Aishwarya Padmakumar, Jesse Thomason, Ayush Shrivastava, Patrick Lange, Anjali
  Narayan-Chen, Spandana Gella, Robinson Piramithu, Gokhan Tur, and Dilek
  Hakkani-Tur. 2021.
\newblock \href {http://arxiv.org/abs/2110.00534} {{TEACh}: Task-driven
  embodied agents that chat}.

\bibitem[{Rogers(2023)}]{Rogers2023-up}
Anna Rogers. 2023.
\newblock Closed {AI} models make bad baselines.
\newblock \url{https://hackingsemantics.xyz/2023/closed-baselines/}.
\newblock Accessed: 2023-4-27.

\bibitem[{Schick et~al.(2023)Schick, Dwivedi-Yu, Dess{\`\i}, Raileanu, Lomeli,
  Zettlemoyer, Cancedda, and Scialom}]{Schick2023-jc}
Timo Schick, Jane Dwivedi-Yu, Roberto Dess{\`\i}, Roberta Raileanu, Maria
  Lomeli, Luke Zettlemoyer, Nicola Cancedda, and Thomas Scialom. 2023.
\newblock \href {http://arxiv.org/abs/2302.04761} {Toolformer: Language models
  can teach themselves to use tools}.

\bibitem[{Schlangen(2003)}]{schlangen2003coherence}
David Schlangen. 2003.
\newblock A coherence-based approach to the interpretation of non-sentential
  utterances in dialogue.

\bibitem[{Shridhar et~al.(2019)Shridhar, Thomason, Gordon, Bisk, Han, Mottaghi,
  Zettlemoyer, and Fox}]{Shridhar2019-sz}
Mohit Shridhar, Jesse Thomason, Daniel Gordon, Yonatan Bisk, Winson Han,
  Roozbeh Mottaghi, Luke Zettlemoyer, and Dieter Fox. 2019.
\newblock \href {http://arxiv.org/abs/1912.01734} {{ALFRED}: A benchmark for
  interpreting grounded instructions for everyday tasks}.

\bibitem[{Sicilia and Alikhani(2022)}]{sicilia2022leather}
Anthony Sicilia and Malihe Alikhani. 2022.
\newblock Leather: A framework for learning to generate human-like text in
  dialogue.
\newblock \emph{arXiv preprint arXiv:2210.07777}.

\bibitem[{Sicilia and Alikhani(2023)}]{sicilia2023learning}
Anthony Sicilia and Malihe Alikhani. 2023.
\newblock Learning to generate equitable text in dialogue from biased training
  data.
\newblock In \emph{Proceedings of the 61st Annual Meeting of the Association
  for Computational Linguistics (Volume 1: Long Papers)}, pages 2898--2917.

\bibitem[{Sicilia et~al.(2023)Sicilia, Asano, Atwell, Cheng, Gupta, Hassan,
  Inan, Nwogu, Sharma, and Alikhani}]{Pittsburgh2023}
Anthony Sicilia, Yuya Asano, Katherine Atwell, Qi~Cheng, Dipunj Gupta, Sabit
  Hassan, Mert Inan, Jennifer Nwogu, Paras Sharma, and Malihe Alikhani. 2023.
\newblock \href
  {https://www.amazon.science/alexa-prize/proceedings/isabel-an-inclusive-and-collaborative-task-oriented-dialogue-system}
  {Isabel: An inclusive and collaborative task-oriented dialogue system}.
\newblock In \emph{Alexa Prize TaskBot Challenge 2 Proceedings}.

\bibitem[{Singh et~al.(2023)Singh, Blukis, Mousavian, Goyal, Xu, Tremblay, Fox,
  Thomason, and Garg}]{Singh2023-ea}
Ishika Singh, Valts Blukis, Arsalan Mousavian, Ankit Goyal, Danfei Xu, Jonathan
  Tremblay, Dieter Fox, Jesse Thomason, and Animesh Garg. 2023.
\newblock {ProgPrompt}: Generating situated robot task plans using large
  language models.
\newblock In \emph{2023 {IEEE} International Conference on Robotics and
  Automation ({ICRA})}, pages 11523--11530.

\bibitem[{Tong et~al.(2024)Tong, Liu, Zhai, Ma, LeCun, and Xie}]{tong2024eyes}
Shengbang Tong, Zhuang Liu, Yuexiang Zhai, Yi~Ma, Yann LeCun, and Saining Xie.
  2024.
\newblock Eyes wide shut? exploring the visual shortcomings of multimodal llms.
\newblock \emph{arXiv preprint arXiv:2401.06209}.

\bibitem[{Vaswani et~al.(2017)Vaswani, Shazeer, Parmar, Uszkoreit, Jones,
  Gomez, Kaiser, and Polosukhin}]{Vaswani2017-kv}
Ashish Vaswani, Noam Shazeer, Niki Parmar, Jakob Uszkoreit, Llion Jones,
  Aidan~N Gomez, {\L}ukasz Kaiser, and Illia Polosukhin. 2017.
\newblock Attention is all you need.
\newblock \emph{Adv. Neural Inf. Process. Syst.}, 30.

\bibitem[{Viegas et~al.(2022)Viegas, Inan, Quandt, and
  Alikhani}]{viegas2022including}
Carla Viegas, Mert Inan, Lorna Quandt, and Malihe Alikhani. 2022.
\newblock Including facial expressions in contextual embeddings for sign
  language generation.
\newblock \emph{arXiv preprint arXiv:2202.05383}.

\bibitem[{Wang et~al.(2018)Wang, Singh, Michael, Hill, Levy, and
  Bowman}]{Wang2018-jg}
Alex Wang, Amanpreet Singh, Julian Michael, Felix Hill, Omer Levy, and Samuel~R
  Bowman. 2018.
\newblock {GLUE}: A multi-task benchmark and analysis platform for natural
  language understanding.
\newblock In \emph{Proceedings of the 2018 {EMNLP} Workshop {BlackboxNLP}:
  Analyzing and Interpreting Neural Networks for {NLP}}, pages 353--355.

\bibitem[{Wang et~al.(2023)Wang, Wang, Lin, Bai, Zhou, Zhou, Wang, and
  Zhou}]{Wang2023-fa}
Peng Wang, Shijie Wang, Junyang Lin, Shuai Bai, Xiaohuan Zhou, Jingren Zhou,
  Xinggang Wang, and Chang Zhou. 2023.
\newblock {ONE-PEACE}: Exploring one general representation model toward
  unlimited modalities.
\newblock \emph{arXiv preprint arXiv:2305. 11172}.

\bibitem[{Warstadt et~al.(2023)Warstadt, Mueller, Choshen, Wilcox, Zhuang,
  Ciro, Mosquera, Williams, Paranjabe, Linzen, and Cotterell}]{Warstadt2023-gv}
Alex Warstadt, Aaron Mueller, Leshem Choshen, Ethan~Gotlieb Wilcox, Chengxu
  Zhuang, Juan Ciro, Rafael Mosquera, Adina Williams, Bhargavi Paranjabe, Tal
  Linzen, and Ryan Cotterell. 2023.
\newblock Findings of the 2023 {B}aby{LM} {C}hallenge: {S}ample-efficient
  pretraining on developmentally plausible corpora.
\newblock In \emph{Proceedings of the 2023 {B}aby{LM} {C}hallenge}. Association
  for Computational Linguistics (ACL).

\bibitem[{Weidinger et~al.(2023)Weidinger, Rauh, Marchal, Manzini, Hendricks,
  Mateos-Garcia, Bergman, Kay, Griffin, Bariach
  et~al.}]{weidinger2023sociotechnical}
Laura Weidinger, Maribeth Rauh, Nahema Marchal, Arianna Manzini, Lisa~Anne
  Hendricks, Juan Mateos-Garcia, Stevie Bergman, Jackie Kay, Conor Griffin, Ben
  Bariach, et~al. 2023.
\newblock Sociotechnical safety evaluation of generative ai systems.
\newblock \emph{arXiv preprint arXiv:2310.11986}.

\bibitem[{Williams and Park(2016)}]{Williams2016-nj}
Jason~D Williams and Menlo Park. 2016.
\newblock The dialog state tracking challenge series : A review.
\newblock \emph{Dialogue \& Discourse}, 7(3):4--33.

\bibitem[{Williams et~al.(2023)Williams, Matuszek, Jokinen, Korpan,
  Pustejovsky, and Scassellati}]{williams2023voice}
Tom Williams, Cynthia Matuszek, Kristiina Jokinen, Raj Korpan, James
  Pustejovsky, and Brian Scassellati. 2023.
\newblock Voice in the machine: Ethical considerations for language-capable
  robots.
\newblock \emph{Communications of the ACM}, 66(8):20--23.

\bibitem[{Williams et~al.(2024{\natexlab{a}})Williams, Matuszek, Mead, and
  DePalma}]{williams2024scarecrows}
Tom Williams, Cynthia Matuszek, Ross Mead, and Nick DePalma.
  2024{\natexlab{a}}.
\newblock Scarecrows in {O}z: The use of large language models in {HRI}.
\newblock \emph{ACM Transactions on Human-Robot Interaction}, 13(1).

\bibitem[{Williams et~al.(2024{\natexlab{b}})Williams, Matuszek, Mead, and
  Depalma}]{Williams2024-ng}
Tom Williams, Cynthia Matuszek, Ross Mead, and Nick Depalma.
  2024{\natexlab{b}}.
\newblock Scarecrows in oz: The use of large language models in {HRI}.
\newblock \emph{J. Hum.-Robot Interact.}, 13(1):1--11.

\bibitem[{Yin et~al.(2021)Yin, Moryossef, Hochgesang, Goldberg, and
  Alikhani}]{yin2021including}
Kayo Yin, Amit Moryossef, Julie Hochgesang, Yoav Goldberg, and Malihe Alikhani.
  2021.
\newblock Including signed languages in natural language processing.
\newblock In \emph{Proceedings of the 59th Annual Meeting of the Association
  for Computational Linguistics and the 11th International Joint Conference on
  Natural Language Processing (Volume 1: Long Papers)}, pages 7347--7360.

\bibitem[{Zhang et~al.(2023)Zhang, Du, Shan, Zhou, Du, Tenenbaum, Shu, and
  Gan}]{Zhang2023-ko}
Hongxin Zhang, Weihua Du, Jiaming Shan, Qinhong Zhou, Yilun Du, Joshua~B
  Tenenbaum, Tianmin Shu, and Chuang Gan. 2023.
\newblock \href {http://arxiv.org/abs/2307.02485} {Building cooperative
  embodied agents modularly with large language models}.

\end{thebibliography}
\balance
\bibliographystyle{acl_natbib}




\end{document}